\documentclass[lettersize,journal]{IEEEtran}
\usepackage{amsmath,amsfonts}
\usepackage{algorithmic}
\usepackage{algorithm}
\usepackage{array}
\usepackage[caption=false,font=normalsize,labelfont=sf,textfont=sf]{subfig}
\usepackage{textcomp}
\usepackage{stfloats}
\usepackage{url}
\usepackage{verbatim}
\usepackage{graphicx}
\usepackage{cite}
\usepackage{ragged2e, pifont}
\usepackage{booktabs,makecell, multirow, tabularx, bbding, amssymb}

\usepackage{booktabs}
\usepackage{indentfirst}
\usepackage{setspace}

\usepackage[pagebackref=true,breaklinks=true,letterpaper=true,colorlinks,bookmarks=false]{hyperref}
\begin{document}

\title{Hierarchical Graph Interaction Transformer with Dynamic Token Clustering for Camouflaged Object Detection}

\author{Siyuan Yao,
        Hao Sun, Tian-Zhu Xiang, Xiao Wang and
        Xiaochun Cao,~\IEEEmembership{Senior Member,~IEEE.}
\thanks{S. Yao and H. Sun are with the School of Computer Science (National Pilot Software Engineering School), Beijing University Of Posts and Telecommunications, Beijing 100876, China. (email: yaosiyuan@bupt.edu.cn; sunhao0504@bupt.edu.cn).}

\thanks{T. Xiang is with the Inception Institute of Artificial Intelligence and G42 Bayanat, Abu Dhabi, UAE. (email: tianzhu.xiang19@gmail.com).}

\thanks{X. Wang is with the School of Software, Beihang University, Beijing 100083, China. (email: xiao$\_$wang@buaa.edu.cn).}

\thanks{X. Cao is with the School of Cyber Science and Technology, Shenzhen Campus, Sun Yat-sen University, Shenzhen 518107, China. (email: caoxiaochun@mail.sysu.edu.cn).}
}

\markboth{Journal of \LaTeX\ Class Files,~Vol.~14, No.~8, August~2021}%
{Shell \MakeLowercase{\textit{et al.}}: A Sample Article Using IEEEtran.cls for IEEE Journals}


\maketitle

\begin{abstract}
Camouflaged object detection (COD) aims to identify the objects that seamlessly blend into the surrounding backgrounds. Due to the intrinsic similarity between the camouflaged objects and the background region, it is extremely challenging to precisely distinguish the camouflaged objects by existing approaches. In this paper, we propose a hierarchical graph interaction network termed HGINet for camouflaged object detection, which is capable of discovering imperceptible objects via effective graph interaction among the hierarchical tokenized features. Specifically, we first design a region-aware token focusing attention (RTFA) with dynamic token clustering to excavate the potentially distinguishable tokens in the local region. Afterwards, a hierarchical graph interaction transformer (HGIT) is proposed to construct bi-directional aligned communication between hierarchical features in the latent interaction space for visual semantics enhancement. Furthermore, we propose a decoder network with confidence aggregated feature fusion (CAFF) modules, which progressively fuses the hierarchical interacted features to refine the local detail in ambiguous regions. Extensive experiments conducted on the prevalent datasets, \textit{i.e.} COD10K, CAMO, NC4K and CHAMELEON demonstrate the superior performance of HGINet compared to existing state-of-the-art methods. Our code is available at https://github.com/Garyson1204/HGINet.
\end{abstract}


\section{Introduction}
\IEEEPARstart{C}AMOUFLAGE is an effective and widespread defensive behavior that causes biological organisms to seamlessly blend into their surroundings, aiming to deceive the perceptual and cognitive system of the predators or prey. As a result, such defensive behavior makes camouflaged objects very challenging to discover for both human and computer perception. The so-called camouflaged object detection (COD) \cite{fan2020camouflaged,mei2021camouflaged,zhai2021mutual,he2023camouflaged} task has gained significant attention in computer vision community with various applications, including medical image segmentation \cite{DBLP:conf/miccai/JiCFCFJS21}, industrial flaw detection \cite{DBLP:conf/iccv/LiuMM0K21} and species discovery \cite{perez2012early}, etc.

\begin{figure}[!t]
\centering
\includegraphics[width=3.5in]{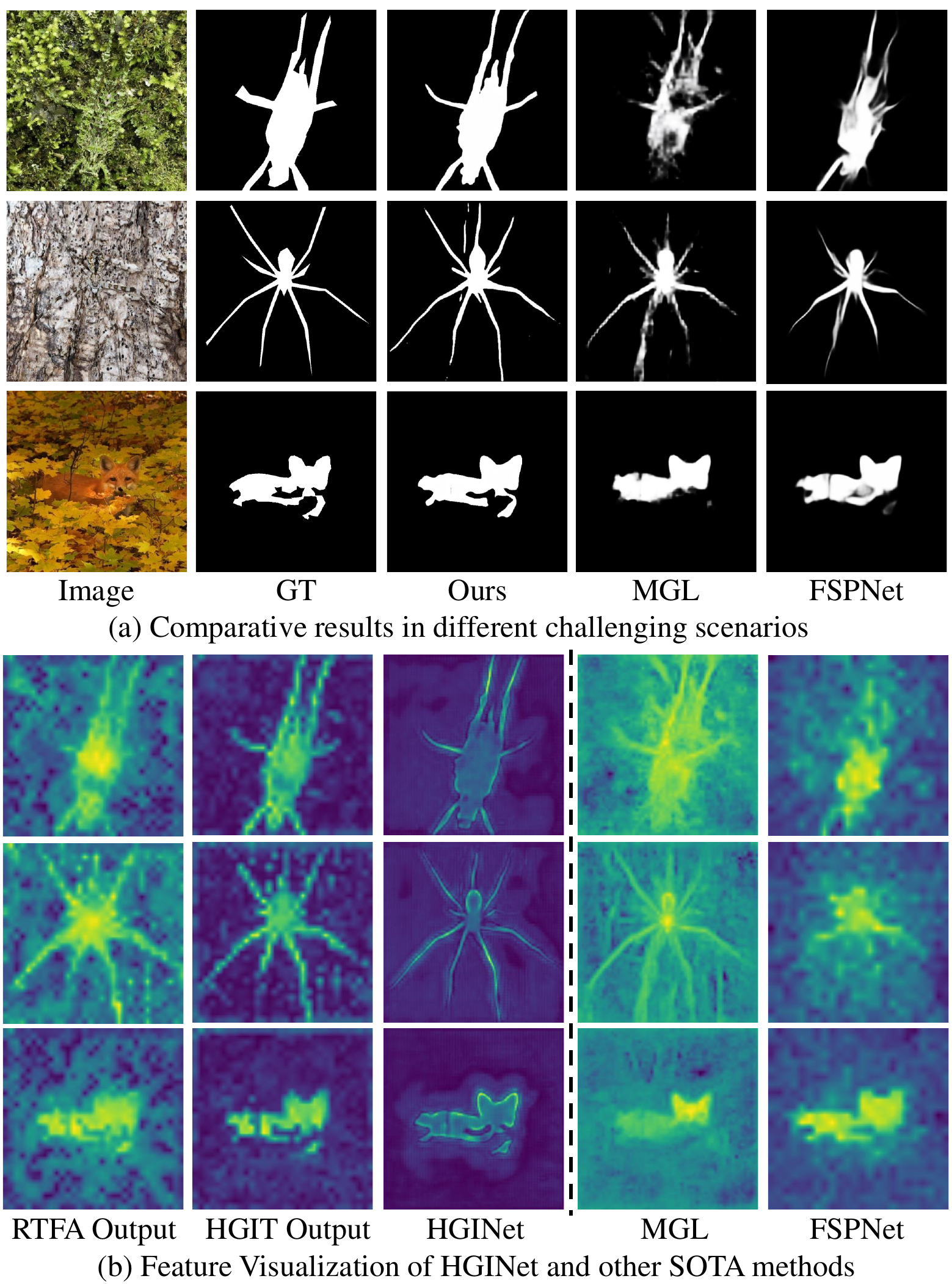}
\caption{Comparison of the prediction results obtained by HGINet with MGL \cite{zhai2021mutual} and FSPNet \cite{huang2023feature}. (a) Visual comparison in different challenging scenarios; (b) The intermediate features learned by HGINet and the SOTA methods.}
\label{fig:illustration}
\end{figure}



Over the past decades, numerous methods \cite{fan2020camouflaged,fan2021concealed,yang2021uncertainty,zhong2022detecting,huang2023feature,he2023camouflaged} have been proposed in this area. The early efforts use handcrafted features, \textit{e.g.}, texture \cite{sengottuvelan2008performance}, motion flow \cite{hou2011detection} and 3D convexity \cite{pan2011study}, while they inevitably suffer drastic performance degradation in complex backgrounds. Recently, deep learning techniques have achieved impressive progress in the COD task. Most of these methods involve developing convolutional neural networks (CNNs) with the strategies of attentive feature distillation \cite{fan2021concealed,DBLP:journals/tcsv/RenHZXXWDH23}, multi-task joint learning \cite{zhai2021mutual,lv2021simultaneously,he2023camouflaged} and bio-inspired vision \cite{DBLP:conf/cvpr/PangZZL20,mei2021camouflaged,pang2022zoom,jia2022segment}, aiming to eliminate the ambiguities caused by the intrinsic similarity between camouflaged object and background regions. However, the CNN based methods rely on stacking convolutional blocks with limited receptive fields, which can not exploit the long-range information, resulting in inferior performance for complex scenarios with heavy occlusions or indistinguishable boundaries. The most recent approaches also employ vision transformer (ViT) \cite{yang2021uncertainty,DBLP:conf/icpr/LiuZTW22,huang2023feature} as encoder backbone or auxiliary module to build long-range relationships between different patch tokens. By recursively merging the neighboring tokens with multi-head self-attention operations, the ViT based approaches enable to capture the beneficial global contextual information and lead the state-of-the-art performance on the COD task by a significant margin.

Despite the demonstrated successes, existing COD techniques potentially suffer from two problems: inconspicuous target-distractor discrimination and insufficient hierarchical semantic interaction. First, since the camouflage strategies are biologically evolved to confuse the predator’s visual system, existing approaches struggle to capture the subtle visual cues for high-quality camouflaged object representation, thus they would be easily misguided by the background distractions. Second, recent advances \cite{sun2021context,lv2021simultaneously,pang2022zoom} show that the hierarchical semantic interaction is critical in the COD task, especially for ambiguous and imperceptible objects. Towards this end, several efforts \cite{zhai2021mutual,huang2023feature} have been made to excavate visual semantics with graph structure to perform hierarchical interaction across the CNN/transformer layers. Though impressive, they fail to construct well-aligned graph representation and lack the long-range perception of the distant graph nodes. As shown in Fig. \ref{fig:illustration} (b), the interaction mechanisms in existing methods like MGL \cite{zhai2021mutual} mutually propagates graph information of the mask branch into the edge branch, which inevitably encounters a significant domain gap due to the discrepancy between these two types of graphs. FSPNet \cite{huang2023feature} pools the ViT features into fixed-size coarse-grained graph nodes via spectral graph convolution, and the pixels with similar features are projected into the same vertex, thus the graph vertices in hierarchical layers have not been carefully aligned and lose the spatial information, leading to inferior segmentation results eventually.

In this paper, we propose a novel COD model termed hierarchical graph interaction network (HGINet), which is capable of discovering the camouflaged object via a hierarchical graph interaction mechanism. Specifically, we first employ a transformer backbone to construct a generic object/background representation, where a region-aware token focusing attention (RTFA) module is designed to enforce the query tokens to focus on the most distinguishable key-value pairs, while the irrelevant tokens are discarded by dynamic token clustering. Afterwards, the visual tokens are reorganized into hierarchical feature maps. Instead of separatively embedding the hierarchical features into graph representation without structural alignment, we embed the feature maps to be latent graph structure and align the graphs using soft-attention. A hierarchical graph interaction transformer (HGIT) is further employed to model the long-range dependencies of the graph nodes, which can be reprojected into the coordinate space. Furthermore, we propose a decoder network with confidence aggregated feature fusion (CAFF) modules, which progressively fuses the hierarchical interacted features to refine the local details in ambiguous regions. Benefiting from the discriminative token excavation and efficient hierarchical feature interaction, HGINet presents high-quality camouflaged object segmentation even in challenging scenarios.

In conclusion, our contributions are summarized below:


$\bullet$ We present a novel COD model termed HGINet, which is capable of discovering the imperceptible objects via effective graph interaction among the hierarchical tokenized features.

$\bullet$ We propose Region-Aware Token Focusing Attention (RTFA) and Hierarchical Graph Interaction Transformer (HGIT) modules, enabling to excavate the potentially distinguishable tokens and construct bi-directional aligned communication to capture the visual semantics.

$\bullet$ We further propose a Confidence Aggregated Feature Fusion (CAFF) decoder to refine the local details in ambiguous regions. Experiments conducted on public COD datasets demonstrate that HGINet achieves superior performance than other state-of-the-art COD methods.

\section{Related Work}
\label{sec:formatting}

\subsection{Camouflaged Object Detection}

Recent years have witnessed significant progress in the field of Camouflaged Object Detection (COD). Generally speaking, existing COD approaches can be roughly divided into three categories: i) feature distillation strategy \cite{mei2021camouflaged,zhong2022detecting,yang2021uncertainty,huang2023feature,10103836}. This strategy attempts to distill the inconspicuous features of camouflaged objects from the background by contextual exploration or introducing additional visual cues. Mei \emph{et~al.} \cite{mei2021camouflaged} leverage contextual information for positioning the target objects and uncertain region refinement. Similarly, Huang \emph{et~al.} \cite{huang2023feature} apply a non-local attention mechanism to process contextual information from the vision transformer based backbone. Besides, Zhong \emph{et~al.} \cite{zhong2022detecting} introduce the frequency domain as an additional clue to better discriminate camouflaged objects from the background.
ii) Multi-task joint learning \cite{zhai2021mutual,sun2022boundary,DBLP:journals/tip/LiYZWZQ22,he2023camouflaged,guo2024cofinet,lv2021simultaneously,DBLP:journals/tip/ZhouZGYZ22,DBLP:journals/tip/ZhaoXZHL23,he2023strategic}. These methods typically utilize auxiliary tasks to segment the target object. For instance, in \cite{DBLP:journals/tip/LiYZWZQ22,he2023camouflaged,he2023strategic}, the boundary-aware priors are introduced to extract features that highlight the structural details of the object. Some works \cite{li2021uncertainty,liu2022modeling} also take the uncertainty of models into consideration to enhance the segmentation reliability. Additionally, PUENet \cite{10159663} respectively models epistemic uncertainty and aleatoric uncertainty for effective segmentation with less model and data bias.
iii) Bio-inspired strategy \cite{9430677,fan2020camouflaged,sun2021context,zhang2022preynet,pang2022zoom,ZoomNeXt}. This approach simulates the behavior process of predators to search and locate camouflaged objects. SINet \cite{fan2020camouflaged} introduces a searching module and an identification module to locate rough areas and detect objects with similar background distractions. ZoomNet \cite{pang2022zoom} imitates human vision by zooming in and out the imperceptible camouflaged objects with mixed scales. Unlike these prior works, our HGINet focuses on capturing discriminative representation of the camouflaged objects and making full use of the hierarchical visual semantics to facilitate the generation of precise segmentation results.

\subsection{Visual Transformer with Token Enhancement}
Transformer is a type of self-attention based neural network that has attracted great interest in the computer vision community. As the pioneering work, Dosovitskiy \emph{et~al.} \cite{DBLP:conf/iclr/DosovitskiyB0WZ21} propose the first vision transformer (ViT) architecture for object recognition and obtain superior performance than the traditional CNNs based approaches. The variants of ViT have been applied to various downstream vision tasks, such as object detection \cite{DBLP:conf/eccv/CarionMSUKZ20}, video understanding \cite{DBLP:conf/cvpr/WangXWSCSX21}, etc. To enhance the visual semantic representation in ViT, some work \cite{DBLP:conf/nips/RaoZLLZH21,DBLP:conf/iclr/LiangGTS0X22,DBLP:conf/cvpr/Tang00XGXT22} attempt to reorganize the visual tokens based on pre-defined scoring mechanism.
DynamicViT \cite{DBLP:conf/nips/RaoZLLZH21} picks up the most important tokens and discards the redundant ones by measuring token-level attention scores. DToP \cite{tang2023dynamic} divides a vision transformer backbone into multiple stages and introduces the auxiliary loss into several stages to measure each token's difficulty. AdaViT\cite{meng2022adavit} adaptively determines the usage of tokens, heads, and layers of vision transformer by a decision network conditioned on input images. ATS \cite{fayyaz2022adaptive} comprehensively selects the most informative tokens exploiting attention scores with the cumulative distribution function.
EViT \cite{DBLP:conf/iclr/LiangGTS0X22} classifies the visual tokens into attentive and inattentive tokens according to class token attention. A-ViT \cite{DBLP:conf/cvpr/YinVAMKM22} introduces distributional prior regularization to perform token pruning, which adaptively adjusts computation cost for different tokens.
But for the COD task, the intrinsic similarity between camouflaged objects and the background makes token aggregation more challenging, which inspires us to propose the region-aware token focusing attention (RTFA) for inconspicuous target-distractor discrimination.

\subsection{Graph Representation Based Methods}
Graph representation based methods have been widely explored in recent years. Among them, Graph Convolution Networks (GCN) are first proposed by \cite{bruna2013spectral} to aggregate features between adjacent nodes. \cite{te2020edge} makes use of first-order approximation of graph convolution to learn the correlation between pixels in 2D features. Moreover, \cite{li2018beyond} learns a graph representation from a 2D feature map through a graph projection, which breaks the barrier between 2D feature maps and graph representations.
To exploit effective message-passing through the structure of transformer, early methods \cite{dwivedi2020generalization,nguyen2022universal,velickovic2017graph} restrict self-attention to local neighborhoods which still suffer from limited expressive power. Recently, global message-passing with transformer architecture has been realized with different approaches and applied to various vision tasks \cite{yuan2023graph, tang2023graph,Ding_2023_ICCV,Li_2022_CVPR,Zhao_2022_CVPR}.
Yuan \emph{et~al.} \cite{yuan2023graph} propose a graph attention transformer network to explore multi-label relationships for image classification which can effectively mine complex inter-label relationships.
TokenGT \cite{kim2022pure} takes all nodes and edges as independent tokens and feeds them to a transformer block to learn the neighbor connection.
Tang \emph{et~al.} \cite{tang2023graph} propose GTGAN to incorporate graph structure into the transformer to capture both local and global relations across nodes in a serial and parallel manner.
Rao \emph{et~al.} \cite{rao2023transg} design a skeleton graph transformer (SGT), which integrates key correlative node features into graph representations.
Ding \emph{et~al.} \cite{Ding_2023_ICCV} utilize an edge-augmented graph transformer to reason the tracking-detection bipartite graph for data association. Different from these methods,
here we convert the feature maps into latent graph space and construct bi-directional aligned communication of the graph nodes.

\begin{figure*}[tb]

\begin{center}
\includegraphics[scale=0.54]{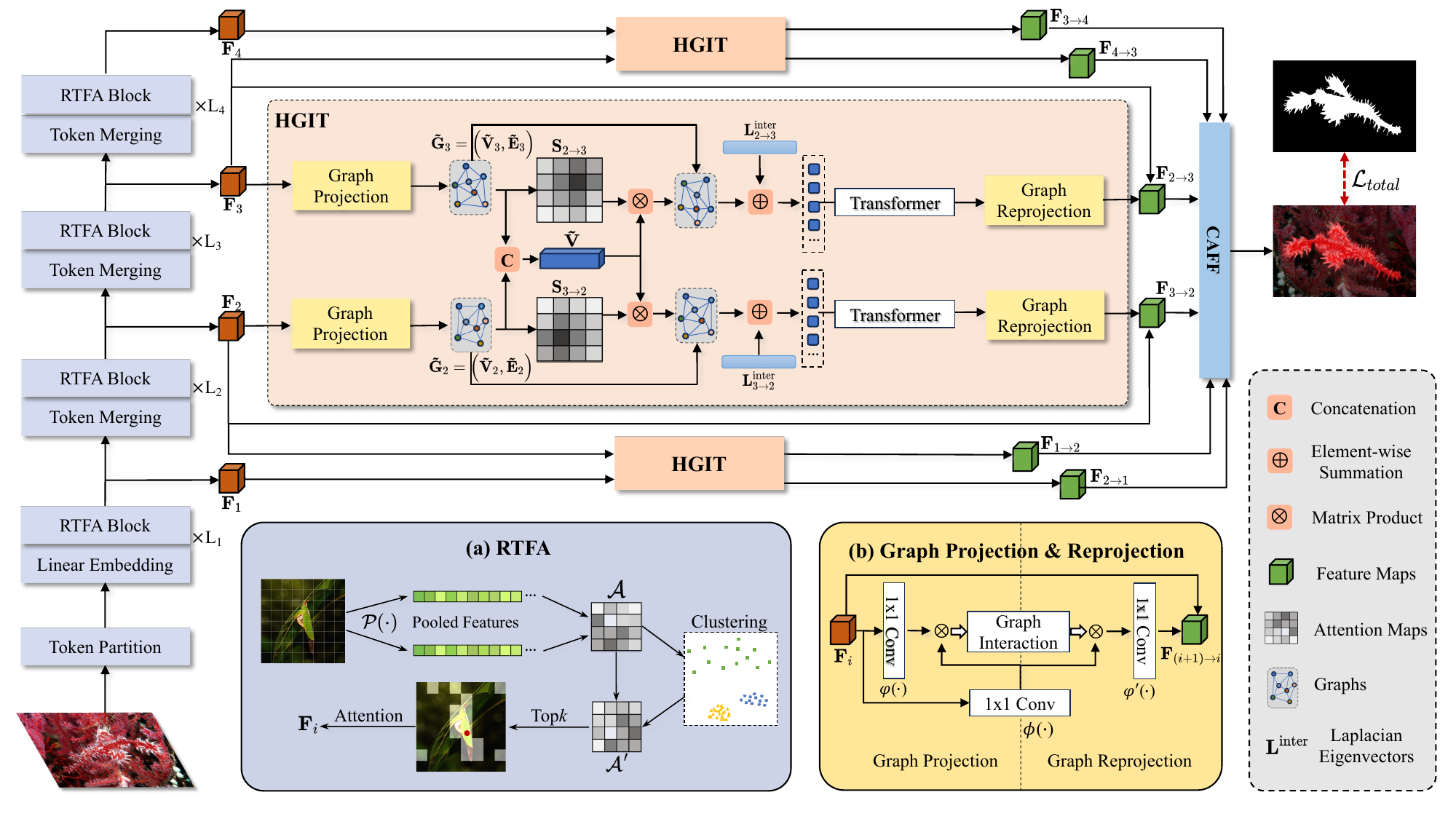}

\end{center}

\caption{The overall architecture of the proposed HGINet. It mainly consists of a transformer backbone with multiple RTFA blocks, a hierarchical graph interaction transformer (HGIT), and a decoder network with confidence aggregated feature fusion (CAFF) modules. (a) illustrates our RTFA, \textit{i.e.}, region-aware token focusing module, which consists of a pooling and dynamic token clustering strategy to excavate the most distinguishable tokens. (b) demonstrates our graph projection and reprojection strategy in latent space.}

\label{fig2}

\end{figure*}

\section{Proposed Method}

\subsection{Overview}

Fig. \ref{fig2} illustrates the overall architecture of our proposed HGINet. The HGINet mainly contains a transformer backbone, a hierarchical graph interaction transformer (HGIT) network, and a decoder network with confidence aggregated feature fusion (CAFF) modules. Specifically, the input image is first fed into a transformer backbone consisting of multiple RTFA blocks, aiming to enforce the query token to attend on the most distinguishable key-value pairs and discard the irrelevant tokens by dynamic token clustering. Then the encoded features in adjacent RTFA blocks are embedded to be latent graph structure to model the semantic relation dependencies between different regions. The constructed graph is passed into a hierarchical graph interaction transformer (HGIT) to enhance the visual semantics. Finally, a decoder network with confidence aggregated feature fusion (CAFF) modules is designed to fuse the hierarchical interacted features for details refinement.

\subsection{Region-aware Token Focusing Attention}
Due to the intrinsic similarity between the camouflaged objects and background surroundings, directly using the self-attention mechanism to establish long-range dependency will inevitably introduce irrelevant interference by the background distractions, resulting in inferior segmentation outputs for camouflaged object discrimination. To address this issue, we propose a region-aware token focusing attention (RTFA) module, allowing the model to excavate the potentially distinguishable tokens using a dynamic token clustering strategy.

Formally, given an input image $\mathbf{I} \in \mathbb{R}^{H\times W \times 3}$ with height $H$ and width $W$, the image is linearly projected and passed through a serial of RTFA blocks to generate the hierarchical feature maps $\mathcal{F} = \left \{ \mathbf{F}_{1},\mathbf{F}_{2},\mathbf{F}_{3},\mathbf{F}_{4} \right \}$. Specifically, suppose the feature size of $\mathbf{F}_{i}$ in the $i$-th RTFA block is $H_{i}\times W_{i}\times C_{i}$, the feature map $\mathbf{F}_{i}$ is divided into $s\times s$ non-overlapped tokens $\mathbf{T}_{i}\in \mathbb{R}^{s^{2} \times \frac{H_{i}W_{i}}{s^{2}}\times C_{i}}$. The query $\mathbf{Q}_{i}$, key $\mathbf{K}_{i}$ and value $\mathbf{V}_{i}$ are acquired by linear projections through matrices $\mathbf{W}^{q}, \mathbf{W}^{k}, \mathbf{W}^{v}\in \mathbb{R} ^{C_{i}\times C_{i}} $, respectively:
\begin{equation}
  \mathbf{Q}_{i}=\mathbf{T}_{i}\mathbf{W}^{q},\; \; \mathbf{K}_{i}=\mathbf{T}_{i}\mathbf{W}^{k},\; \; \mathbf{V}_{i}=\mathbf{T}_{i}\mathbf{W}^{v}.
  \label{eq:eq1}
\end{equation}

Afterwards, we construct a region-aware affinity matrix, which indicates how much the tokens are semantically related. We perform average pooling on the query token $\mathbf{Q}_{i}$ and key token $\mathbf{K}_{i}$ into $\hat{\mathbf{Q}}_{i}, \hat{\mathbf{K}}_{i}\in \mathbb{R} ^{s^{2} \times \left ( 1\times C_{i}\right )}$ to squeeze the spatial redundancy in local region. Then we calculate the affinity matrix $\mathcal{A}\in \mathbb{R} ^{s^{2} \times s^{2}} $ by multiplying $\hat{\mathbf{Q}}_{i}$ and the transpose of $\hat{\mathbf{K}}_{i}$ to measure the semantic associations between the query-key tokens, which can be formulated as:
\begin{equation}
   \mathcal{A}=\hat{\mathbf{Q}}_{i}\left(\hat{\mathbf{K}}_{i}\right)^{T} = \mathcal{P}\left(\mathbf{Q}_{i}\right) \left(\mathcal{P}\left(\mathbf{K}_{i}\right)\right)^{T},
  \label{eq:eq2}
\end{equation}
where $\mathcal{P}\left(\cdot\right)$ denotes the average pooling operation. Then we utilize a dynamic token clustering method based on DPC-KNN \cite{DBLP:journals/kbs/DuDJ16} to discard the token redundancy. We calculate the local density $\rho_{p}$ of each visual token in affinity matrix $\mathcal{A}$ based on its k-nearest neighbors:

\begin{equation}
\rho_{p} = \exp\left ( -\frac{1}{k}\sum_{\mathbf{a}_{q}\in \mathrm{KNN}(\mathbf{a}_{p})} \left \| \mathbf{a}_{p}-\mathbf{a}_{q} \right \|_{2}^{2} ) \right ),
  \label{eq:eq3}
\end{equation}
where $\mathbf{a}_{p}$ and $\mathbf{a}_{q}$ denotes the visual tokens with index $p$ and $q$ in $\mathcal{A}$, $\mathrm{KNN}(\mathbf{a}_{p})$ denotes the k-nearest neighbors of $\mathbf{a}_{p}$. For each token, a distance indicator $\delta_{p}$ is utilized to measure the closeness between this token and the surrounding tokens as follows:
\begin{equation}
\delta_{p}=\left\{\begin{matrix}
 \min \limits_{q:\rho_{p}> \rho_{q}}\left \| \mathbf{a}_{p}-\mathbf{a}_{q} \right \|_{2}^{2}, && \text{if $\exists q \hspace{3pt} \mbox{s.t.} \hspace{3pt} \mathbf{a}_{p}>\mathbf{a}_{q} $} \\
 \max \limits_{q:\rho_{p}> \rho_{q}}\left \| \mathbf{a}_{p}-\mathbf{a}_{q} \right \|_{2}^{2}, && \text{otherwise}
\end{matrix}\right.
\end{equation}
where $\delta_{p}$ measures the distance between the token $\mathbf{a}_{p}$ to the nearest region with higher local density. We calculate the score of each token by $\rho_{p} \times \delta_{p}$. The higher scores indicate that the token $\mathbf{a}_{p}$ presents informative visual semantics for camouflaged object discrimination. The cluster centers with the top $k$ scores are selected to construct the discriminative clustered token $\mathbf{C}_{i}\in \mathbb{R} ^{s^{2} \times k}$, which is further concatenated with the key-value pairs for token enhancement:
\begin{equation}
  \mathbf{K}_{i}^{\prime}=\mathrm{Concat}\left ( \mathbf{K}_{i},\mathbf{C}_{i} \right ), \; \;
  \mathbf{V}_{i}^{\prime}=\mathrm{Concat}\left ( \mathbf{V}_{i},\mathbf{C}_{i} \right ),
  \label{eq:eq4}
\end{equation}
where $\mathbf{K}_{i}^{\prime}$ and $\mathbf{V}_{i}^{\prime}$ denote the enhanced key-value pairs. Finally, we apply multi-head self-attention within those enhanced tokens:
\begin{equation}
  \mathbf{F}_{i}=\mathrm{SA}\left ( \mathbf{Q}_{i},\mathbf{K}_{i}^{\prime}, \mathbf{V}_{i}^{\prime} \right ).
  \label{eq:eq5}
\end{equation}

\subsection{Hierarchical Graph Interaction Transformer}
Although RTFA is capable to construct discriminative camouflaged object representation by dynamically suppressing the irrelevant key-value tokens, it still lacks a hierarchical interaction mechanism to exchange the visual semantics in multiple RTFA blocks. Keep this in mind, we embed the hierarchical feature maps to be latent graph structures and align them using soft-attention. A hierarchical graph interaction transformer (HGIT) is employed to exchange the graph information in adjacent blocks, which assists the model in constructing long-range dependencies of the graph nodes for bi-directional message communication.
\noindent\textbf{Latent Space Graph Projection.}
For the feature map $\mathbf{F}_{i}$, we firstly use bilinear interpolation to unify the feature size to be $H'\times W' \times C'$ in multiple RTFA blocks and reshape as $ \mathbf{F}_{i}' \in \mathbb{R} ^{ L \times C'}$, where $C'$ is the unified channel dimension and $L=H'\times W'$ denotes the spatial resolution. Then the reshaped feature $ \mathbf{F}_{i}'$ is passed through two $1 \times 1$ convolution layers, \textit{i.e.} $\phi ( \cdot)\in \mathbb{R} ^{L \times N}$ and $\varphi(\cdot) \in \mathbb{R} ^{L \times C'}$ to reduce the dimension and aggregate feature in different channels, respectively. A latent graph $\mathbf{\tilde{G}}_{i} = \left ( \mathbf{\tilde{V}}_{i}, \mathbf{\tilde{E}}_{i} \right )$ can be constructed by:

\begin{equation}
    \mathbf{\tilde{V}}_{i} = \phi\left ( \mathbf{F}_{i}'\right )^{T}\varphi(\mathbf{F}_{i}'),
  \label{eq:eq6}
\end{equation}
where $\mathbf{\tilde{V}}_{i}\in \mathbb{R} ^{ N\times C'}$ is the graph nodes. $N$ denotes the pre-defined vertices number. The inner-product output in Eq. \ref{eq:eq6} can be regarded as a projection operation, which performs matrix multiplication along the spatial coordinate's dimension to generate feature map $\mathbf{\tilde{V}}_{i}$ in latent interaction spaces. By doing this, the original features map $\mathbf{F}_{i}$ in the coordinate space are embedded to more compact representation $\mathbf{\tilde{V}}_{i}$, making the identification of the regions with similar visual semantics at different locations to be more consistent in the latent interaction spaces.


\noindent\textbf{Hierarchical Graph Interaction.}
After obtaining graph representations via latent space graph projection, we apply a simple yet effective interaction approach to create the local alignment and communications between the graphs in hierarchical transformer blocks. Specifically, for the graph nodes $\mathbf{\tilde{V}}_{i}$ in $i$-th stage and $\mathbf{\tilde{V}}_{i+1}$ in $(i+1)$-th stage, we use the non-local operation \cite{DBLP:conf/cvpr/0004GGH18} with softmax to perform bi-directional interaction:

\begin{equation}
  \begin{aligned}
    &\mathbf{S}_{i\to (i+1)} = \mathrm{Softmax}(\psi_{1}(\mathbf{\tilde{V}}_{i})\theta_{1}^{T}(\mathbf{\tilde{V}}_{i+1})),\\
    &\mathbf{S}_{(i+1) \to i} = \mathrm{Softmax}(\psi_{2}(\mathbf{\tilde{V}}_{i+1})\theta_{2}^{T}(\mathbf{\tilde{V}}_{i})),
  \label{eq:eq7}
  \end{aligned}
\end{equation}
where $\mathbf{S}_{i\to (i+1)} \in \mathbb{R} ^{N\times N}$ denotes the interaction from $\mathbf{\tilde{V}}_{i}$ to $\mathbf{\tilde{V}}_{i+1}$ and vice versa for $\mathbf{S}_{(i+1) \to i}$.  The function $\psi_{1}(\cdot)$, $\psi_{2}(\cdot)$, $\theta_{1}(\cdot)$ and $\theta_{2}(\cdot)$ are learnable transformations on the graph nodes. $\mathbf{S}_{i\to (i+1)}$ and $\mathbf{S}_{(i+1)\to i}$ can be regarded as the alignment matrices measuring the correlation between the nodes in dual graphs, which hint the complementary visual semantics corresponding to the hierarchical feature maps.

Meanwhile, we concatenate the graph $\mathbf{\tilde{V}}_{i}, \mathbf{\tilde{V}}_{i+1}$ and squeeze the feature channel into $\mathbf{\tilde{V}} \in \mathbb{R} ^{N\times C'}$ to combine both graph information. Then we perform graph interaction by multiplying $\mathbf{\tilde{V}}$ with $\mathbf{S}_{i\to (i+1)}$ and $\mathbf{S}_{(i+1) \to i}$, respectively:

\begin{equation}
  \begin{aligned}
    &\mathbf{\tilde{V}}_{i}^{\mathrm{inter}} = \mathbf{S}_{(i+1) \to i}\mathbf{\tilde{V}} + \mathbf{\tilde{V}}_{i},\\
    &\mathbf{\tilde{V}}_{i+1}^{\mathrm{inter}} = \mathbf{S}_{i\to (i+1)}\mathbf{\tilde{V}} + \mathbf{\tilde{V}}_{i+1},
  \label{eq:eq8}
  \end{aligned}
\end{equation}
where $\mathbf{\tilde{V}}_{i}^{\mathrm{inter}}, \mathbf{\tilde{V}}_{i+1}^{\mathrm{inter}} \in \mathbb{R} ^{N\times C'}$ denote the graph nodes aggregated with bi-directional interaction. By performing such interaction, the latent graph nodes $\mathbf{\tilde{V}}_{i}$ and $\mathbf{\tilde{V}}_{i+1}$ are simultaneously enhanced, leading to more powerful visual semantic mining of the camouflaged objects.

\noindent\textbf{Transformer Learning and Reprojection.}
We further employ a transformer to model the high-order dependencies of the interacted graph nodes $\mathbf{\tilde{V}}_{i}^{\mathrm{inter}}$ and $\mathbf{\tilde{V}}_{i+1}^{\mathrm{inter}}$. For simplicity, here we take $\mathbf{\tilde{V}}_{i}^{\mathrm{inter}}$ as example, the inner product of $\mathbf{\tilde{V}}_{i}^{\mathrm{inter}}$ is computed to generate the adjacent matrix $\mathbf{A}$ and the corresponding diagonal degree matrix $\mathbf{D}$. The graph information is incorporated into the transformer block by taking the Laplacian matrix as positional embedding:
\begin{equation}
    \mathbf{L}_{(i+1)\to i}^{\mathrm{inter}} = \mathbf{I} - \mathbf{D}^{-1/2}\mathbf{A}\mathbf{D}^{-1/2} ,
  \label{eq:eq10}
\end{equation}
where $\mathbf{L}_{(i+1)\to i}^{\mathrm{inter}}$ is the Laplacian matrix of interacted graph node $\mathbf{\tilde{V}}_{i}^{\mathrm{inter}}$. After that, $\mathbf{\tilde{V}}_{i}^{\mathrm{inter}}$ and $\mathbf{L}_{(i+1)\to i}^{\mathrm{inter}}$ are passed into a standard transformer block with multi-head self-attention:

\begin{equation}
    \mathbf{\tilde{V}}_{i}' =\mathrm{MHSA}\left ( \mathbf{\tilde{V}}_{i}^{\mathrm{inter}} + \mathbf{L}_{(i+1)\to i}^{\mathrm{inter}} \right ),
  \label{eq:eq12}
\end{equation}
where $\mathbf{\tilde{V}}_{i}' \in \mathbb{R} ^{N \times C'}$ is the enhanced graph node. It is then reprojected back into the original coordinate by $\mathbf{F}_{i}^{\mathrm{Reproj}}=\phi\left ( \mathbf{F}_{i}'\right )\mathbf{\tilde{V}}_{i}'$, thus the spatial structure can be recovered. Finally, we utilize a function $\varphi'\left ( \cdot \right )$ with $1\times 1$ convolution to combine the reprojected visual semantics $\mathbf{F}_{i}^{\mathrm{Reproj}}$ with the original feature map $\mathbf{F}_{i}$ by:
\begin{equation}
    \mathbf{F}_{(i+1)\to i} =\varphi'(\mathbf{F}_{i}^{\mathrm{Reproj}})+\mathbf{F}_{i},
  \label{eq:eq13}
\end{equation}
where $\mathbf{F}_{(i+1)\to i}$ is the enhanced feature map via latent graph interaction from $(i+1)$-th to $i$-th transformer block.
\begin{figure}[!t]
\centering
\setlength{\abovecaptionskip}{0.7mm}
\includegraphics[width=3.53in]{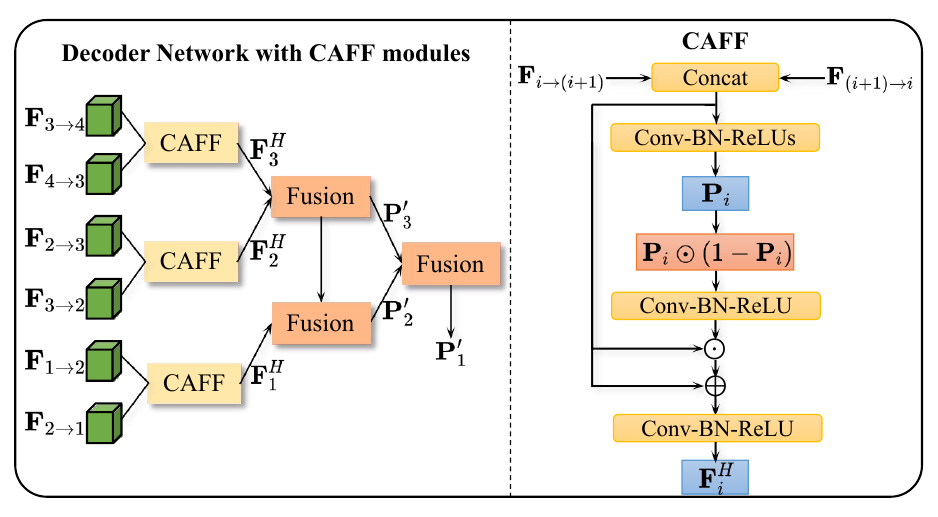}
\caption{Details of our decoder network with CAFF modules. ``Fusion" in the decoder network consists of a Conv-BN-ReLU layer and Pixel Shuffle. }
\label{fig:fig3}
\end{figure}

\begin{table*}[t]
  \footnotesize
  \centering
  \caption{Quantitative comparison with 20 SOTA methods on three COD benchmark datasets. Notes $\uparrow/\downarrow$ denotes the higher/lower the better, and the best and second best are \textbf{bolded} and \underline{underlined} for highlighting, respectively. }
    \begin{tabular}{l|c|cccc|cccc|cccc}
    \toprule
    \multirow{2}[3]{*}{\textbf{Methods}} & \multirow{2}[3]{*}{\textbf{Year}} & \multicolumn{4}{c}{\textbf{CAMO (250)}}      & \multicolumn{4}{c}{\textbf{COD10K (2,026)}}    & \multicolumn{4}{c}{\textbf{NC4K (4,121)}} \\
\cmidrule{3-14} & & $S_{\alpha}\uparrow $ &  $F_{\beta}^{w}\uparrow $  & $E_{\phi}^{m}\uparrow $ & $\mathcal{M}\downarrow$ & $S_{\alpha}\uparrow $ &  $F_{\beta}^{w}\uparrow $  &   $E_{\phi}^{m}\uparrow $   & $\mathcal{M}\downarrow$ & $S_{\alpha}\uparrow $ &  $F_{\beta}^{w}\uparrow $  &   $E_{\phi}^{m}\uparrow $  & $\mathcal{M}\downarrow$ \\
    \midrule
    SINet\cite{fan2020camouflaged}  & CVPR'20 & 0.751  &   0.606  &  0.771  &  0.100  & 0.771  &  0.551 & 0.806  & 0.051  &  0.808  &  0.723 &  0.871 &  0.058  \\
    SLSR\cite{lv2021simultaneously}  & CVPR'21 & 0.787    &  0.696  &  0.838  &  0.080  & 0.804  &  0.673 & 0.880  &  0.037  &   0.840    & 0.766   &  0.895  &  0.048 \\
    PFNet\cite{mei2021camouflaged} & CVPR'21 & 0.782  &    0.695  &  0.842  &  0.085  & 0.800  &  0.660 & 0.877  &  0.040  &  0.829  &    0.745  &   0.888   &  0.053  \\
    MGL-R\cite{zhai2021mutual} & CVPR'21 & 0.775   &  0.673  &  0.812  &  0.088  & 0.814  &  0.666 & 0.852  &   0.035  &    0.833   &   0.740    &    0.867   &    0.052   \\
    UJSC\cite{li2021uncertainty}  & CVPR'21 & 0.800   &  0.728  &  0.859  &  0.073  & 0.809  &  0.684 & 0.884  & 0.035  &  0.842  &   0.771 & 0.898 &  0.047 \\
    C2FNet\cite{sun2021context} & IJCAI'21 & 0.796  &  0.719  &  0.854  &  0.080  & 0.813  &  0.686 & 0.890  &   0.036  &  0.838  &  0.762  &  0.897  &  0.049 \\
    UGTR\cite{yang2021uncertainty}  & ICCV'21 & 0.784   &  0.684  &  0.822  &  0.086  & 0.817  &  0.666 & 0.853  & 0.036  &  0.839 & 0.747  &  0.875  &  0.052 \\
    FindNet\cite{DBLP:journals/tip/LiYZWZQ22} & TIP'22  & 0.794   &  0.717  &  0.851  &  0.079  & 0.818  &  0.699 & 0.891  &   0.034  &  0.841  &  0.771  & 0.897  &  0.048  \\
    OCENet\cite{liu2022modeling} & WACV'22 & 0.802   &  0.723  &  0.852  &  0.080  & 0.827  & 0.707  &  0.894 & 0.033  &  0.853  &   0.785  & 0.903  &  0.045 \\
    BGNet\cite{sun2022boundary} & IJCAI'22 & 0.812   &  0.749  &  0.870  &  0.073  & 0.831  &  0.722 &  0.901   &  0.033  &  0.851  &  0.788 &  0.907  &  0.044 \\
    SegMaR\cite{jia2022segment} & CVPR'22 & 0.815  &  0.753  &  0.874  &  0.071  & 0.833  &  0.724 & 0.899  & 0.034  &  0.841  &   0.781  &   0.896  &  0.046  \\
    ZoomNet\cite{pang2022zoom} & CVPR'22 & 0.820  &  0.752  &  0.878  &  0.066  & 0.838  &  0.729 & 0.888  & 0.029  &  0.853  &  0.784  &   0.896   &  0.043 \\
    SINet-v2\cite{fan2021concealed} & TPAMI'22 & 0.820  &  0.743  &  0.882  &  0.070  & 0.815  &  0.680 & 0.887  &  0.037  &  0.847  &  0.770  &  0.903  &  0.048 \\
    HitNet\cite{hu2023high} & AAAI'23 & 0.849  &  0.809  &  0.906  &   0.055  &  0.871  &  \underline{0.806} & \underline{0.935}  & 0.023  &  0.875  &  0.834  &  0.926  &  0.037 \\
    PENet\cite{ijcai2023p124} & IJCAI'23 & 0.828  &  0.771  &  0.890  &   0.063  & 0.831  &  0.723 & 0.908  & 0.031  &  0.855  &  0.795  &  0.912  &  0.042 \\
    FEDER\cite{he2023camouflaged} & CVPR'23 & 0.836  &  0.807  &  0.897  &   0.066  & 0.844  &  0.748 & 0.911  & 0.029  &  0.862  &  0.824  &  0.913  &  0.042 \\
    FSPNet\cite{huang2023feature} & CVPR'23 & 0.856   &  0.799  &  0.899  &  0.050  & 0.851  &  0.735 & 0.895  &  0.026  &  0.879  &  0.816  &  0.915  &  0.035  \\
    PUENet\cite{10159663} & TIP'23 & 0.860   &  0.821  &  0.918  &  0.050  & 0.868  &  0.792  & 0.934  &  0.023  &  \underline{0.892}  &  \underline{0.853}  &  \underline{0.941}  &  \underline{0.030} \\
    PopNet\cite{wu2023source} & ICCV'23 &  0.808  &  0.744  &  0.859  &  0.077  &  0.851  &  0.757  & 0.910  &   0.028  &  0.861  &  0.802  &  0.909  &  0.042  \\
    CamoFocus\cite{Khan_2024_WACV} & WACV'24 &  \underline{0.873}  &  \underline{0.842}  &  \underline{0.926}  &  \underline{0.043}  &  \underline{0.873}  &  0.802  & \underline{0.935}  &   \underline{0.021}  &  0.889  &  \underline{0.853}  &  0.936  &  \underline{0.030}  \\
    \midrule
    \textbf{HGINet (Ours)} &   -   & \textbf{0.874}  &   \textbf{0.848}  &  \textbf{0.937}  &  \textbf{0.041}  & \textbf{0.882}  &  \textbf{0.821} & \textbf{0.949}  & \textbf{0.019}  & \textbf{0.894}  & \textbf{ 0.865} &  \textbf{0.947} &  \textbf{0.027}  \\
    \bottomrule
    \end{tabular}%
  \label{tab:tab1}%
\end{table*}%
\subsection{Decoder Network}
To effectively fuse the hierarchical features, we design a decoder network with confidence aggregated feature fusion (CAFF) modules to refine the local details in ambiguous regions, see Fig. \ref{fig:fig3}. Concretely, we first concatenate the feature maps $\mathbf{F}_{(i+1)\to i}$, $\mathbf{F}_{i \to (i+1)}$ and squeeze them to be $\mathbf{F}_{i}^{C} \in \mathbb{R}^{H_{i}\times W_{i}\times C_{i}}$ in single interaction stage, then the concatenate feature $\mathbf{F}_{i}^{C}$ is fused using a set of convolutional blocks to obtain a coarse-grained prediction $\mathbf{P}_{i}$:
\begin{equation}
  \begin{aligned}
    \mathbf{P}_{i}&= \mathrm{Sigmoid}\left (\mathrm{Conv}\left ( \mathbf{F}_{i}^{C} \right )\right ) \\
    & =  \mathrm{Sigmoid}\left (\mathrm{Conv}\left (\mathrm{Concat}\left ( \mathbf{F}_{(i+1)\to i},\mathbf{F}_{i \to (i+1)} \right )\right )\right ),
  \label{eq:eq14}
  \end{aligned}
\end{equation}
where $\mathrm{Conv}\left (\cdot \right )$ stands for the Conv-BN-ReLU layers, and $\mathrm{Sigmoid}\left (\cdot \right )$ represents the Sigmoid function. Note that such coarse-grained prediction may induce unsatisfying results in ambiguous regions, as the per-pixel prediction score in these regions are generally unconfident. To alleviate this issue, we highlight the ambiguous regions by multiplying $\mathbf{P}_{i}$ and $1 - \mathbf{P}_{i}$ to reweight the concatenated feature for refinement:
\begin{equation}
    \mathbf{F}_{i}^{H} = \mathbf{F}_{i}^{C} \odot \mathrm{Conv} \left (\mathbf{P}_{i}\odot \left (1 - \mathbf{P}_{i} \right )\right ) + \mathbf{F}_{i}^{C},
  \label{eq:eq15}
\end{equation}
where $\odot$ represents the element-wise multiplication. The term $\mathbf{P}_{i}\odot \left (1 - \mathbf{P}_{i} \right )$ can be regarded as a pixel-wise ambiguity indicator. If $\mathbf{P}_{i}$ is close to 0.5, $\mathbf{P}_{i}\odot \left (1 - \mathbf{P}_{i} \right )$ would be large; otherwise if $\mathbf{P}_{i}$ is close to 0 or 1, $\mathbf{P}_{i}\odot \left (1 - \mathbf{P}_{i} \right )$ would be tiny. By multiplying this term with $\mathbf{F}_{i}^{C}$ using skip connection, the ambiguous boundary regions can be aggregated, while the high confident regions are also preserved. Then we employ fusion blocks consist of Conv-BN-ReLU layers and Sigmoid function to obtain the refined prediction result:
\begin{equation}
\mathbf{P}'_{i} = \begin{cases}\mathrm{Sigmoid}\left (\mathrm{Conv}\left ( \mathrm{Concat}\left (\mathbf{F}_{i}^{*}, \mathbf{F}_{i+1}^{*}\right ) \right )\right ), i = 1
 \\\mathrm{Sigmoid}\left (\mathrm{Conv}\left ( \mathrm{Concat}\left (\mathbf{F}_{i}^{*}, \mathbf{F}_{i-1}^{H}\right ) \right )\right ),i = 2,3
\end{cases},
  \label{eq:eq16}
\end{equation}
where $\mathbf{P}'_{i}$ denotes the refined prediction output for the $i$-th fusion block, the intermediate feature $\mathbf{F}_{i}^{*}$ is given by: 
\begin{equation}
\mathbf{F}_{i}^{*} = \begin{cases}\mathrm{Conv}\left ( \mathrm{Concat}\left (\mathbf{F}_{i+1}^{*}, \mathbf{F}_{i}^{H}\right )\right ), i = 1,2
 \\\mathbf{F}_{3}^{H},i=3
\end{cases}.
  \label{eq:eq17}
\end{equation}

We employ the weighted binary cross entropy loss and weighted IoU loss to train the model, which can be given by:
\begin{equation}
  \begin{aligned}
    \mathcal{L}_{total} &= \sum_{i=1}^{3}  2^{(i-3)} \lambda \mathcal{L}_{wbce}\left ( \mathbf{P}'_{i}, \mathbf{G} \right ) \\
    &+ \sum_{i=1}^{3}  2^{(i-3)} (1-\lambda) \mathcal{L}_{wiou} \left ( \mathbf{P}'_{i}, \mathbf{G} \right ),
  \label{eq:eq18}
  \end{aligned}
\end{equation}
where $\mathbf{G}$ denotes the ground truth of the camouflaged images, $i$ denotes the interaction stage in our decoder. $\mathcal{L}_{wbce}$ and $\mathcal{L}_{wiou}$ are weighted BCE loss and weighted IoU loss, respectively, $\lambda$ is the weight coefficient to balance the influence of two loss functions.


\begin{figure*}

\centering
\setlength{\abovecaptionskip}{-1mm}
\includegraphics[scale=0.47]{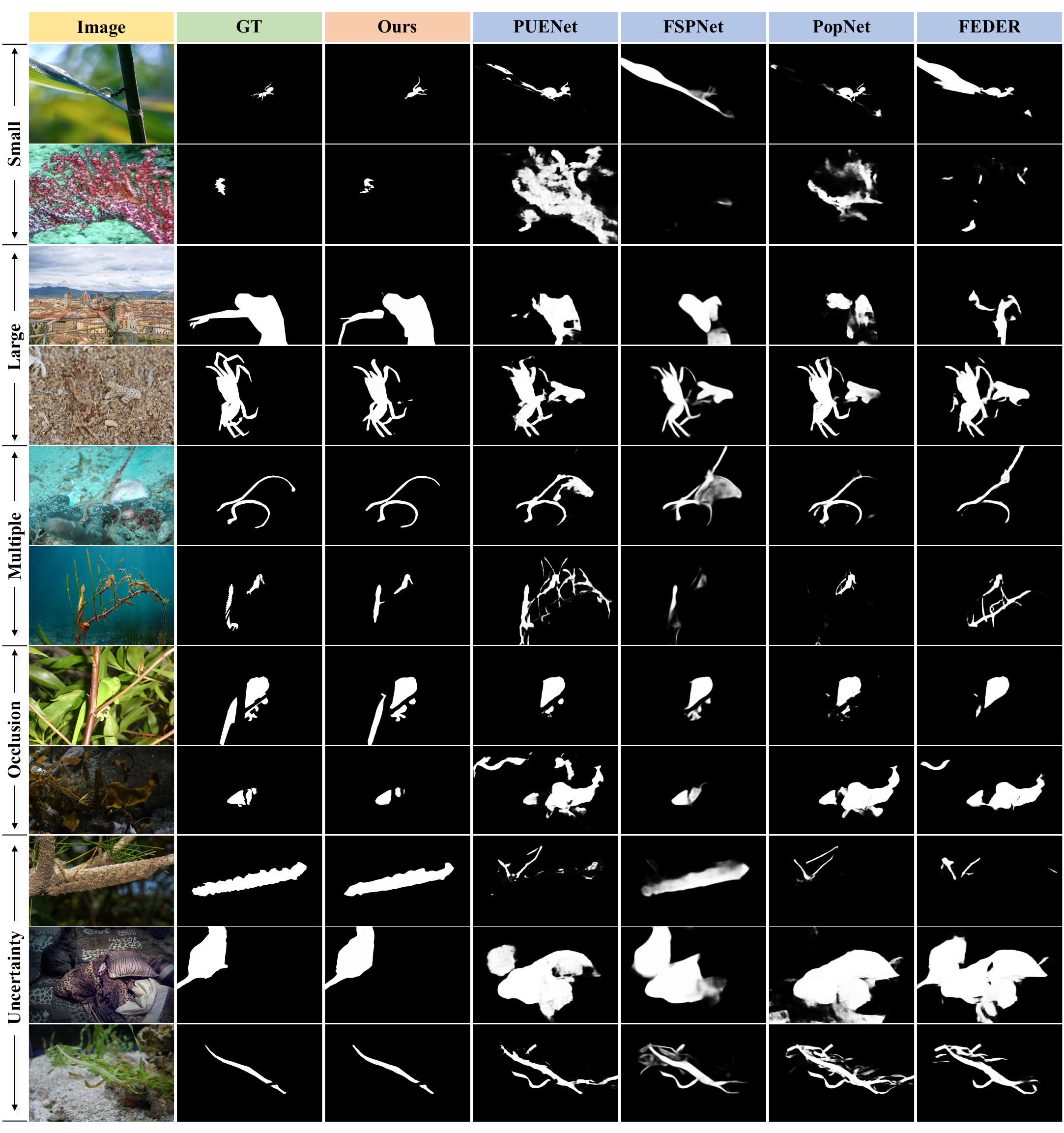}

\caption{Visual comparison with several representative state-of-the-art methods in challenging scenarios, including small, large, multiple, occluded objects, and confused boundaries with great uncertainty. Please zoom in for details.}
\label{fig:fig4}
\end{figure*}

\begin{table}[t]
  \footnotesize
  \centering
  \caption{Quantitative comparison with 14 SOTA methods on CHAMELEON dataset. Notes $\uparrow/\downarrow$ denotes the higher/lower the better, and the best and second best are \textbf{bolded} and \underline{underlined} for highlighting, respectively. }
    \begin{tabular}{l|c|cccc}
    \toprule
    \multirow{2}[3]{*}{\textbf{Methods}} & \multirow{2}[3]{*}{\textbf{Year}} & \multicolumn{4}{c}{\textbf{CHAMELEON (76)}} \\
\cmidrule{3-6} & & $S_{\alpha}\uparrow $ &  $F_{\beta}^{w}\uparrow $  & $E_{\phi}^{m}\uparrow $ & $\mathcal{M}\downarrow$ \\
    \midrule
    SINet\cite{fan2020camouflaged}  & CVPR'20 & 0.872  &   0.827  &  0.936  &  0.034 \\
    PFNet\cite{mei2021camouflaged} & CVPR'21 & 0.882  &    0.826  &  0.922  &  0.033   \\
    MGL-R\cite{zhai2021mutual} & CVPR'21 & 0.893   &  0.834  &  0.918  &  0.030  \\
    UJSC\cite{li2021uncertainty}  & CVPR'21 & 0.894   &  0.848  &  0.943  &  0.030  \\
    UGTR\cite{yang2021uncertainty}  & ICCV'21 & 0.888   &  0.796  &  0.918  &  0.031  \\
    FindNet\cite{DBLP:journals/tip/LiYZWZQ22} & TIP'22  & 0.895   &  0.858  &  0.946  &  0.027  \\
    BGNet\cite{sun2022boundary} & IJCAI'22 & 0.901   &  0.860  &  0.943  &  0.027  \\
    SegMaR\cite{jia2022segment} & CVPR'22 & 0.906  &  \underline{0.872}  &  0.951  &  0.025   \\
    ZoomNet\cite{pang2022zoom} & CVPR'22 & 0.902  &  0.864  &  0.943  &  0.023  \\
    SINet-v2\cite{fan2021concealed} & TPAMI'22 & 0.888  &  0.835  &  0.942  &  0.030  \\
    PENet\cite{ijcai2023p124} & IJCAI'23 & 0.902  &  0.851  &  \underline{0.960}  &   0.024  \\
    FEDER\cite{he2023camouflaged} & CVPR'23 & 0.887  &  0.836  &  0.949  &   0.029  \\
    FSPNet\cite{huang2023feature} & CVPR'23 & 0.907   &  0.848  &  0.941  &  0.024  \\
    PUENet\cite{10159663} & TIP'23 & \underline{0.910}   &  0.869  &  0.957  &  \underline{0.022} \\
    \midrule
    \textbf{HGINet (Ours)} &   \textendash   & \textbf{0.915}  &  \textbf{0.889}  &  \textbf{0.970}  &  \textbf{0.018} \\
    \bottomrule
    \end{tabular}%
  \label{tab:tab6}%
\end{table}%

\section{Experiments and Results}

\subsection{Experiment Settings}

\noindent {\bf Datasets.} We train and evaluate the proposed HGINet on four widely used COD datasets, \emph{i.e.}, CAMO \cite{le2019anabranch}, COD10K \cite{fan2020camouflaged}, NC4K \cite{lv2021simultaneously}, and CHAMELEON. CAMO is the first proposed COD dataset that contains 1,250 camouflaged images, which splits 1,000 images for training and others for testing. COD10K is the largest COD benchmark, which contains 3,040 training images and 2,026 testing images. NC4K is a recently released COD benchmark which includes 4,121 images for testing. Following the standard benchmark, we take 3,040 training images from COD10K and 1,000 training images from CAMO for model training, whereas the rest images form the testing datasets. CHAMELEON is the smallest test dataset that contains 76 high-resolution images with hand-annotated pixel-level labels.

\begin{table*}[t]
  \footnotesize
  \centering
  \caption{Ablation studies of our HGINet variants. The results are reported on CAMO, COD10K, and NC4K datasets for quantitative analysis.}
    \begin{tabular}{ccc|cccc|cccc|cccc}
    \toprule
    \multicolumn{3}{c}{\textbf{Configurations}} & \multicolumn{4}{c}{\textbf{CAMO (250)}}      & \multicolumn{4}{c}{\textbf{COD10K (2,026)}}    & \multicolumn{4}{c}{\textbf{NC4K (4,121)}} \\
\cmidrule{1-15} \textbf{RTFA} & \textbf{HGIT} & \textbf{CAFF} &   $S_{\alpha}\uparrow $ &  $F_{\beta}^{w}\uparrow $  & $E_{\phi}^{m}\uparrow $ & $\mathcal{M}\downarrow$ & $S_{\alpha}\uparrow $ &  $F_{\beta}^{w}\uparrow $  &   $E_{\phi}^{m}\uparrow $   & $\mathcal{M}\downarrow$ & $S_{\alpha}\uparrow $ &  $F_{\beta}^{w}\uparrow $  &   $E_{\phi}^{m}\uparrow $  & $\mathcal{M}\downarrow$ \\
    \midrule
      &  &  &  0.824  &   0.778  &  0.884  &  0.055  & 0.817  &  0.722 &  0.875  & 0.027  &  0.859  &  0.812 &  0.909 & 0.037 \\
    \CheckmarkBold  &  &  &  0.868  &   0.797  &  0.892  &  0.053  & 0.874  &  0.759 &  0.886  & 0.025  &  0.881  &  0.801 &  0.893 &  0.037  \\
    \CheckmarkBold  & \CheckmarkBold &  &  0.862  &   0.832  &  0.928  &  0.046  & 0.875  &  0.809  & 0.942  &  0.020  &  0.886  &  0.853 &  0.941 &  0.030  \\
    \CheckmarkBold  &  & \CheckmarkBold &  0.861  &   0.819  &  0.912  &  0.048  & 0.872  &  0.776 & 0.910  & 0.023  &  0.885  &  0.833 &  0.925 &  0.033  \\
    \CheckmarkBold  & \CheckmarkBold & \CheckmarkBold &  \textbf{0.874}  &  \textbf{0.848}  &  \textbf{0.937}  &  \textbf{0.041}  &  \textbf{0.882}  & \textbf{0.821}   &  \textbf{0.949} &  \textbf{0.019}  &  \textbf{0.894}  & \textbf{0.865}   &  \textbf{0.947} &  \textbf{0.027}  \\
    \bottomrule
    \end{tabular}%
  \label{tab:tab2}%
\end{table*}%

\noindent {\bf Evaluation Metrics.} Four commonly used evaluation metrics are applied for the COD task, including S-measure ($S_{\alpha}$) \cite{Cheng2021sMeasure}, weighted F-measure ($F_{\beta}^{w}$) \cite{6909433}, mean E-measure ($E_{\phi}^{m}$) \cite{ijcai2018p97} and mean absolute error ($\mathcal{M}$). Specifically, $S_{\alpha}$ evaluates both region-aware and object-aware structural similarities. The region-aware structural similarity aims to capture the structure information of object parts, whereas object-aware structural similarity is designed to capture the structure information of the complete foreground objects. $F_{\beta}^{w}$ is proposed based on the methodology of the F-measure. It replaces the precision and recall with weighted precision and weighted recall to measure the exactness and completeness, respectively. E-measure ($E_{\phi}$) is proposed for measuring binary foreground maps. It combines pixel-level matching and image-level statistics based on human visual perception. Following PUENet, we adopt mean E-measure ($E_{\phi}^{m}$) as our metrics for fair comparison.

\noindent {\bf Implementation Details.} The proposed method is implemented by PyTorch on 4 NVIDIA GeForce RTX 3090 GPUs and optimized by Adam. The learning rate is initialized to 1e-4 and then decreased by a factor of 10 in every 50 epochs. We take the transformer network in \cite{DBLP:conf/cvpr/ZhuWKZL23} pre-trained on ImageNet as the backbone, while other modules are randomly initialized. The top $k$ parameters are set to $1, 4, 16, 64$ at the 4 stages. In HGIT, we project the features into graph with 8 vertices and we use 2 transformer layers with 8 heads to model the long-range dependencies. Images are resized to 512\(\times\)512 both in the training and inference phases. The hyper parameter $\lambda$ is set to 0.7. During training, the batch size is set to 8 and the model is trained with 200 epochs.

\subsection{Comparison with State-of-the-arts}


\noindent {\bf Quantitative Comparison.}
We compare our HGINet with 20 state-of-the-art COD methods on three widely used COD datasets, \emph{i.e.}, CAMO, COD10K, and NC4K, and show the quantitative performance in Tab. \ref{tab:tab1}. For fair comparison, predictions of the above methods are directly generated by well-trained models released by authors. It is clear that our HGINet outperforms all methods on these datasets. Compared to the recently proposed state-of-the-art PUENet \cite{10159663}, our proposed method obtain average performance gains of 1.2\%, 2.8\%, 1.4\%, and 15.1\% on for $S_{\alpha} $,  $F_{\beta}^{w} $, $E_{\phi}^{m} $, and $\mathcal{M}$ respectively on these three datasets. Compared to the recently proposed transformer-based FSPNet \cite{huang2023feature}, our proposed method gains average performance progress of 2.5\%, 7.9\%, 4.6\%, and 18.7\% in terms of $S_{\alpha} $,  $F_{\beta}^{w} $, $E_{\phi}^{m} $, and $\mathcal{M}$. Meanwhile, it is worth noting that COD10K and NC4K are the two most challenging datasets in terms of the number of images and the segmentation difficulty. On COD10K and NC4K, our HGIT averagely surpasses PUENet with 6.1\% and 3.1\% performance gains and FSPNet with 12.1\% and 8.5\% over four metrics on COD10K and NC4K, respectively. Besides, we take 14 state-of-the-art COD methods to compare with our HGINet on the CHAMELEON benchmark, see Tab. \ref{tab:tab6}. Our HGINet still outperforms all methods and gains an average performance progress of 5.6\% over four metrics toward PUENet. Benefiting from the discriminative token excavation and hierarchical feature interaction/fusion of RTFA, HGIT, and CAFF, the proposed HGINet achieves significant performance improvement in all datasets.
\\ { \bf Qualitative Comparison.}
As shown in Fig. \ref{fig:fig4}, we perform the visual comparison of the proposed HGINet with the SOTA methods in several typical scenarios, including small, large, multiple, occlusion objects, and confused boundaries with great uncertainty. Compared with other SOTA methods, our model is more discriminative for locating and segmenting small objects from the background and has a better overall perception of large camouflaged objects. Our model also achieves better results within the occlusion and multi-object segmentation problems. In the case of boundary confusion, which is denoted as ``Uncertainty", our HGINet can preserve more local details compared to other methods and it is capable to distinguish the subtle discrepancies between camouflaged objects and interference backgrounds. For example, in the 4-th row, our model can accurately distinguish the crab disguised on the beach. The 3-rd and 7-th rows also reveal that owing to the effectiveness of HGIT, the objects extracted by HGINet are more complete.

\begin{table}[t]
  \footnotesize
  \centering
  \setlength{\tabcolsep}{3.3pt}
  \caption{Ablation studies on region-aware token focusing (RTFA) module. The results are reported on CAMO and COD10K datasets for quantitative analysis. w/ Pooling, w/ Clustering denote baseline with pooling and both clustering and dynamic token clustering strategy, respectively.}
    \begin{tabular}{l|cccc|cccc}
    \toprule
    \multirow{2}[3]{*}{\textbf{Settings}} & \multicolumn{4}{c}{\textbf{CAMO (250)}}      & \multicolumn{4}{c}{\textbf{COD10K (2,026)}}   \\
\cmidrule{2-9} &  $S_{\alpha}\uparrow $ &  $F_{\beta}^{w}\uparrow $  & $E_{\phi}^{m}\uparrow $ & $\mathcal{M}\downarrow$ & $S_{\alpha}\uparrow $ &  $F_{\beta}^{w}\uparrow $  &   $E_{\phi}^{m}\uparrow $   & $\mathcal{M}\downarrow$ \\
    \midrule
    Baseline  & 0.852  &  0.814  &  0.912  &  0.051  & 0.867   & 0.797  &  0.939 & 0.023  \\
    w/ Pooling  & 0.861   &  0.830  &  0.923  &  0.047  &  0.874   & 0.809  &  0.941 & 0.021 \\
    w/ Clustering  & \textbf{0.874}  &  \textbf{0.848}  &  \textbf{0.937}  &  \textbf{0.041}  &  \textbf{0.882}  & \textbf{0.821}   &  \textbf{0.949} &  \textbf{0.019}   \\
    \bottomrule
    \end{tabular}%
  \label{tab:tab3}%
\end{table}%

\subsection{Ablation Study}
\noindent\textbf{Ablation of Different Variants.}
To verify the effectiveness of each module in our HGINet, we selectively choose our proposed RTFA, HGIT, and CAFF modules for comparison. Among them, if RTFA is not used in the experiment, it is replaced by a vanilla vision transformer, HGIT is replaced by a set of Conv-BN-ReLU layers, and CAFF is displaced by the classical FPN head. The results are shown in Tab. \ref{tab:tab2}. Take the results in CAMO as an example, the vanilla vision transformer achieves $0.778$ and $0.884$ in terms of weighted F-measure and E-measure metrics. Since our RTFA employs pooling and clustering to excavate the potentially distinguishable tokens, the variant approach with only RTFA module can obtain improvements with $2.4\%$ and $0.9\%$. By adding HGIT, the weighted F-measure and E-measure obtain obvious improvement ratios of $4.4\%$ and $4.0\%$, respectively. If the CAFF module is further introduced into the whole network, the model further outperforms $1.9\%$ and $1.0\%$, respectively. The results conducted on other two datasets also reveal the similar performance changes. Such results verify that the proposed RTFA, HGIT, and CAFF modules can significantly improve the performance on different COD datasets.
\begin{figure*}[!t]
\setlength{\abovecaptionskip}{-4mm}
\setlength{\belowcaptionskip}{-4mm}

\centering
\includegraphics[width=7in]{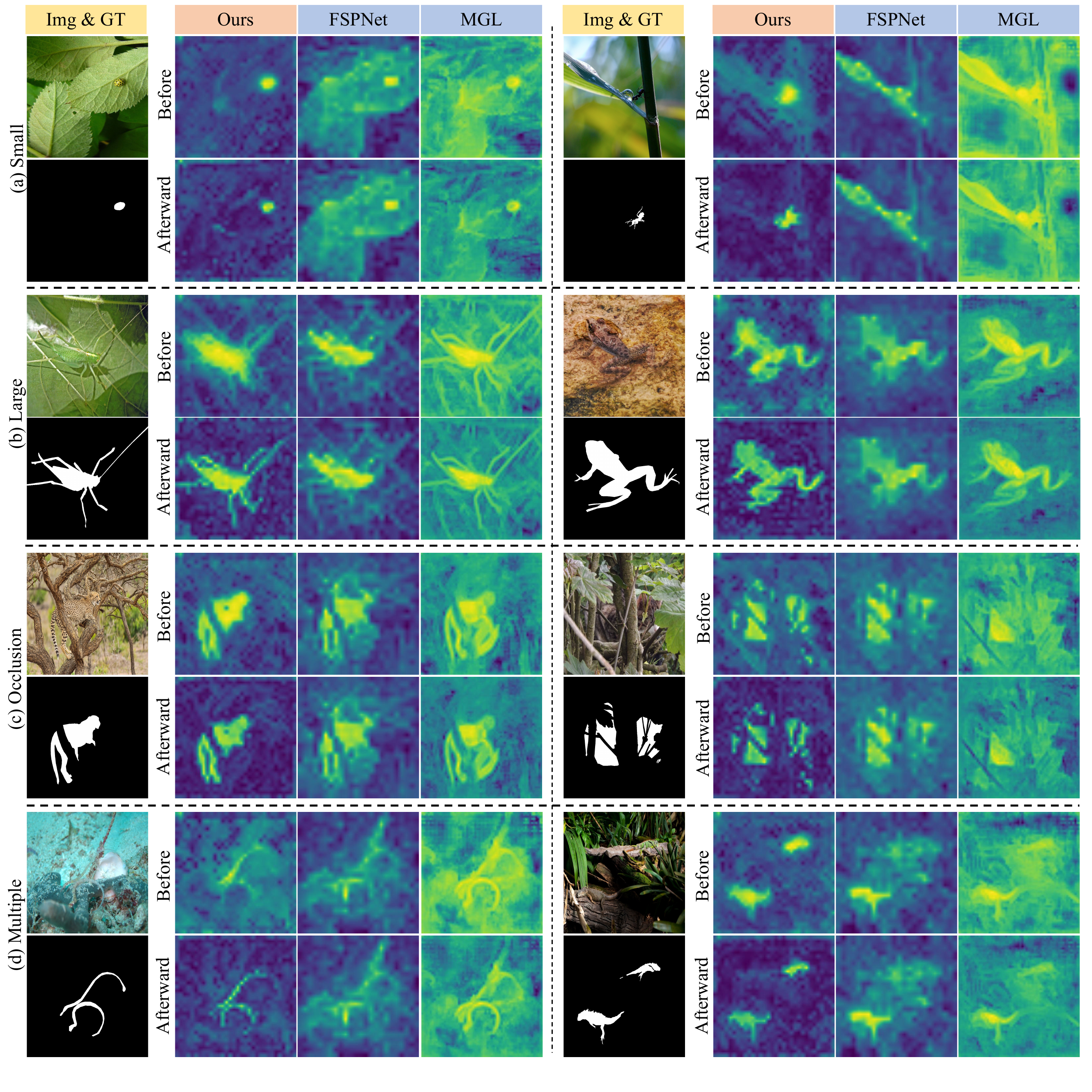}
\caption{Visualization results of features before and after passing the HGIT module in HGINet. Features before and after passing the graph convolution networks in FSPNet\cite{huang2023feature} and MGL\cite{zhai2021mutual} are also presented for comparison. Please zoom in for details.}
\label{fig:fig5}
\end{figure*}
\begin{table}[t]
  \footnotesize
  \centering
  \setlength{\tabcolsep}{3.3pt}
  \caption{Ablation studies on different graph projection strategies. The results are reported on CAMO and COD10K datasets for quantitative analysis.}
    \begin{tabular}{c|cccc|cccc}
    \toprule
    \multirow{2}[3]{*}{\textbf{Settings}} & \multicolumn{4}{c}{\textbf{CAMO (250)}}      & \multicolumn{4}{c}{\textbf{COD10K (2,026)}}   \\
\cmidrule{2-9} &  $S_{\alpha}\uparrow $ &  $F_{\beta}^{w}\uparrow $  & $E_{\phi}^{m}\uparrow $ & $\mathcal{M}\downarrow$ & $S_{\alpha}\uparrow $ &  $F_{\beta}^{w}\uparrow $  &   $E_{\phi}^{m}\uparrow $   & $\mathcal{M}\downarrow$ \\
    \midrule
    \textit{Spatial}  &   0.856  &  0.824  &  0.922  &  0.049  & 0.875   & 0.811  &  0.944 & 0.020  \\
    \textit{PixCluster}  & \textbf{0.874}   &  0.821  &  0.913  &  0.049  &  0.880   & 0.779  &  0.915 & 0.023 \\
    \textit{Latent}  & \textbf{0.874}  &  \textbf{0.848}  &  \textbf{0.937}  &  \textbf{0.041}  &  \textbf{0.882}  & \textbf{0.821}   &  \textbf{0.949} &  \textbf{0.019} \\
    \bottomrule
    \end{tabular}%
  \label{tab:tab4}%
\end{table}%
\begin{table}[t]
  \footnotesize
  \centering
  \setlength{\tabcolsep}{3.3pt}
  \caption{Ablation studies on HGIT. The results are reported on CAMO and COD10K datasets for quantitative analysis.}
    \begin{tabular}{c|cccc|cccc}
    \toprule
    \multirow{2}[3]{*}{\textbf{Settings}} & \multicolumn{4}{c}{\textbf{CAMO (250)}}      & \multicolumn{4}{c}{\textbf{COD10K (2,026)}}   \\
\cmidrule{2-9} &  $S_{\alpha}\uparrow $ &  $F_{\beta}^{w}\uparrow $  & $E_{\phi}^{m}\uparrow $ & $\mathcal{M}\downarrow$ & $S_{\alpha}\uparrow $ &  $F_{\beta}^{w}\uparrow $  &   $E_{\phi}^{m}\uparrow $   & $\mathcal{M}\downarrow$ \\
    \midrule
    w/o HGIT  &   0.861  &   0.819  &  0.912  &  0.048  & 0.872  &  0.776 & 0.910  & 0.023  \\
    1HGIT & 0.865   &  0.834  &  0.930  &  0.047  &  0.875   & 0.809  &  0.943 & 0.021 \\
    2HGIT & 0.858   &  0.826  &  0.922  &  0.049  &  0.870   & 0.802  &  0.942 & 0.021 \\
    3HGIT  & \textbf{0.874}  &  \textbf{0.848}  &  \textbf{0.937}  &  \textbf{0.041}  &  \textbf{0.882}  & \textbf{0.821}   &  \textbf{0.949} &  \textbf{0.019} \\
    \bottomrule
    \end{tabular}%
  \label{tab:tab5}%
\end{table}%
\begin{table}[t]
  \footnotesize
  \centering
  \setlength{\tabcolsep}{3.3pt}
  \caption{Ablation studies of our decoder network with CAFFs. The results are reported on CAMO and COD10K datasets for quantitative analysis.}
    \begin{tabular}{c|cccc|cccc}
    \toprule
    \multirow{2}[3]{*}{\textbf{Settings}} & \multicolumn{4}{c}{\textbf{CAMO (250)}}      & \multicolumn{4}{c}{\textbf{COD10K (2,026)}} \\
\cmidrule{2-9} &   $S_{\alpha}\uparrow $ &  $F_{\beta}^{w}\uparrow $  & $E_{\phi}^{m}\uparrow $ & $\mathcal{M}\downarrow$ & $S_{\alpha}\uparrow $ &  $F_{\beta}^{w}\uparrow $  &   $E_{\phi}^{m}\uparrow $   & $\mathcal{M}\downarrow$  \\
    \midrule
    FPN   &  0.862  &   0.832  &  0.928  &  0.046  & 0.875  &  0.809 &  0.942  & 0.020   \\
    AIMs  &  0.864  &   0.831  &  0.931  &  0.045  & 0.874  &  0.809  & 0.941  &  0.020   \\
    CAFFs & \textbf{0.874}  &  \textbf{0.848}  &  \textbf{0.937}  &  \textbf{0.041}  &  \textbf{0.882}  & \textbf{0.821}   &  \textbf{0.949} &  \textbf{0.019}\\
    \bottomrule
    \end{tabular}%
  \label{tab:tab11}%
\end{table}%
\begin{figure*}[t]
\setlength{\abovecaptionskip}{1mm}
\centering
\includegraphics[width=7in]{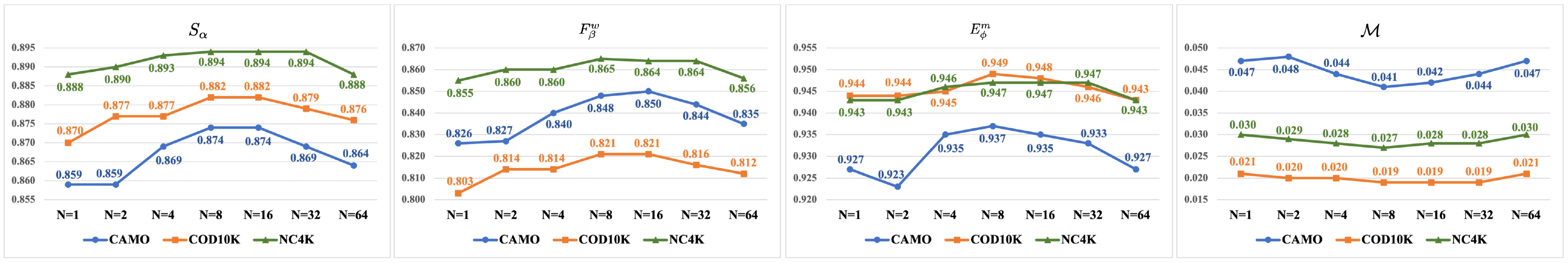}
\caption{Ablation studies of the number of graph node settings on HGIT. The results are reported on CAMO, COD10K, and NC4K datasets for quantitative analysis. Note that the model performance is better when the values of $S_{\alpha}$, $F_{\beta}^{w}$, and $E_{\phi}^{m}$ are larger and on the contrary, the value of $\mathcal{M}$ is smaller.}
\label{fig:fig6}
\end{figure*}

\noindent\textbf{Effectiveness of RTFA.} To verify the effects of our pooling and dynamic token clustering strategy within RTFA, we conduct an ablation analysis to test the variants with or without these two operations. We establish a model without pooling and dynamic token clustering strategy as the baseline, ``w/ Pooling"  refers to introducing the pooling operation into the baseline, and ``w/ Clustering" refers to pooling and dynamic clustering. We demonstrate the results in Tab. \ref{tab:tab3}. Compared with the baseline, the experimental results show that the model with pooling acquires more discrimination capability by perceiving region awareness and discarding the irrelevant tokens, while the ablation method with dynamic token clustering obtains further performance improvement compared with the ``w/ Pooling" version.

\begin{table*}[t]
  \footnotesize
  \centering
  \caption{Ablation studies of the hyper-parameter settings of the transformer architecture within HGIT. All of the variations set graph vertices number $N=32$ for fair comparison. }
    \begin{tabular}{cc|cccc|cccc|cccc}
    \toprule
    \multicolumn{2}{c}{\multirow{2}[3]{*}{\textbf{Settings}}} & \multicolumn{4}{c}{\textbf{CAMO (250)}}      & \multicolumn{4}{c}{\textbf{COD10K (2,026)}}    & \multicolumn{4}{c}{\textbf{NC4K (4,121)}} \\
\cmidrule{3-14}  &  &   $S_{\alpha}\uparrow $ &  $F_{\beta}^{w}\uparrow $  & $E_{\phi}^{m}\uparrow $ & $\mathcal{M}\downarrow$ & $S_{\alpha}\uparrow $ &  $F_{\beta}^{w}\uparrow $  &   $E_{\phi}^{m}\uparrow $   & $\mathcal{M}\downarrow$ & $S_{\alpha}\uparrow $ &  $F_{\beta}^{w}\uparrow $  &   $E_{\phi}^{m}\uparrow $  & $\mathcal{M}\downarrow$ \\
    \midrule
    \textit{$l=0$}  & \textit{$/$} &  0.853  &   0.814  &  0.919  &  0.052  & 0.871  &  0.802 &  0.940  & 0.022  &  0.886  &  0.850 &  0.940 &  0.030  \\
    \textit{$l=1$}  & \textit{$h=8$} &  0.862  &   0.832  &  0.927  &  0.047  & 0.877  &  0.814 &  0.946  & 0.020  &  0.890  &  0.859 &  0.945 &  0.029  \\
    \textit{$l=2$}  & \textit{$h=8$}  &  0.869  &   \textbf{0.844}  &  0.933  &  \textbf{0.044}  & 0.879  &  0.816  & 0.946  &  \textbf{0.019}  &  0.894  &  0.864 &  0.947 &  0.028  \\
    \textit{$l=4$}  & \textit{$h=8$} &  \textbf{0.870}  &   \textbf{0.844}  &  \textbf{0.935}  &  \textbf{0.044}  & \textbf{0.881}  &  \textbf{0.819} & \textbf{0.948}  & \textbf{0.019}  &  \textbf{0.896}  &  \textbf{0.865} &  \textbf{0.948} &  \textbf{0.027}  \\
    \textit{$l=2$}  & \textit{$h=4$} &   0.861  &   0.829  &  0.924  &  0.047  & 0.876  &  0.814 & 0.944  & 0.020  &  0.887  &  0.854 &  0.942 &  0.030   \\
    \textit{$l=2$}  & \textit{$h=16$} &  0.861  &   0.830  &  0.925  &  0.047  & 0.876  &  0.813 & 0.942  & 0.020  &  0.888  &  0.856 &  0.943 &  0.029  \\
    \bottomrule
    \end{tabular}%
  \label{tab:tab9}%
\end{table*}%

\noindent\textbf{Effectiveness of HGIT Module.} We conduct an ablation study to investigate whether different graph projection methods in HGIT will affect the segmentation performance, the results are shown in Tab. \ref{tab:tab4}. Among them, ``\textit{Spatial}" projects features into a spatial space by a $1\times1$ convolution and then aggregates global and local features in spatial space; ``\textit{PixCluster}" adopts a learnable clustering method proposed in \cite{li2018beyond} to learn cluster centers which bring similar features together and separate different features; ``\textit{Latent}" represents our latent space graph projection method. Results show that our projection method in latent space is more effective in modeling the dependencies within the graph structure, as this strategy enables better exchanges of complementary visual
semantics via efficient bi-directional aligned communication among the graph nodes without losing the spatial information.

Besides, we also explore the effect of changing the number of HGIT modules in the proposed network, which is shown in Tab. \ref{tab:tab5}. In our settings, the ``1HGIT" model constructs single HGIT module between the backbone features $\left \{ \mathbf{F}_{2},\mathbf{F}_{3}\right )$. The ``2HGIT" model constructs two HGIT modules and performs graph interaction between the backbone features $\left \{ \mathbf{F}_{1},\mathbf{F}_{2}\right )$ and $\left \{ \mathbf{F}_{3},\mathbf{F}_{4}\right )$. ``3HGIT" model adopts three HGIT modules between $\left \{ \mathbf{F}_{1},\mathbf{F}_{2}\right )$, $\left \{ \mathbf{F}_{2},\mathbf{F}_{3}\right )$ and $\left \{ \mathbf{F}_{3},\mathbf{F}_{4}\right )$. Additionally, ``w/o HGIT" denotes that the hierarchical graph interaction modules are removed. The experimental results show that the hierarchical graph interaction leads to obvious performance gains, the ``3HGIT" which performs feature interaction within every adjacent RTFA block obtains the best results. Compare with ``w/o HGIT", ``1HGIT" gains 3.6\% improvement in term of $E_{\phi}^{m}$ on COD10K. The ``3HGIT" gains further improvement with 0.6\%, indicating that the hierarchical graph interaction in multiple transformer blocks presents beneficial message communication for camouflaged object discovery. Besides, we also notice that ``2HGIT" does not achieve performance improvement compared with ``1HGIT", such result indicates that among these RTFA blocks, the features between the intermediate second and third blocks provide more informative semantics for camouflaged object discrimination.

\noindent\textbf{Comparative Visualization of HGIT.} To further illustrate the effect of our HGIT, in Fig. \ref{fig:fig5}, we visualize the feature maps before and after passing the HGIT module, and compare them with the corresponding feature maps in FSPNet \cite{huang2023feature} and MGL \cite{zhai2021mutual} in four challenging scenarios, including (a) small, (b) large, (c) occluded, and (d) multiple objects. The comparative results show that before passing the graph modules, the RTFA block can effectively distinguish the camouflaged objects using the pooling and dynamic token clustering strategy, but the background regions still present unignorable interference which hampers the segmentation results. Besides, the visualization results after HGIT indicates that our module can effectively eliminate some ambiguous areas, owing to the effectiveness of the hierarchical aligned graph interaction. Compared to FSPNet and MGL, the visualization results show that our method can promote structural details of objects due to the long-range dependencies modeling by HGIT.

\noindent\textbf{Effectiveness of CAFFs.} We also compare our CAFFs with different decoder heads to illustrate the effectiveness of our decoder network with CAFF modules. Here we replace CAFFs with an FPN head and AIMs proposed in FSPNet \cite{huang2023feature}, the experimental results are presented in Tab. \ref{tab:tab11}. Owing to the confidence aggregation mechanism which effectively refines the segmentation outputs in ambiguous regions, the CAFFs can preserve high-quality local details and achieve higher performance than AIM and FPN heads.

\noindent\textbf{Hyper-parameter Settings of HGIT.} In order to explore the hyper-parameter settings of our HGIT module, we conduct two sets of ablation experiments on HGIT. As shown in Tab. \ref{tab:tab9}, we quantitatively analyze the influence of the number of transformer layers $l$ and MHSA heads $h$ in HGIT with the graph vertices number 32. The results show that our model achieves the best performance when $l=4$ and $h=8$. However, adding more transformer layers inevitably increases the computation cost, so we set $l=2$ and $h=8$ to balance the performance and computation efficiency. In addition, we investigate the performance variations by changing the number $N$ of graph vertices in Fig. \ref{fig:fig6}, where we set $N=1, 2, 4, 8, 16, 32, 64$ respectively. The performance improves when $N$ increases from $1$ to $8$, while increasing larger number of graph vertices after $N=8$ actually deteriorates the performance. Therefore, for COD task, setting the graph vertices to $8$ is more suitable to model the non-local relationships between the camouflaged object and background regions.

\noindent\textbf{Generalization Ability for Unseen Scenes.} To investigate the performance of our HGINet and other COD models in unseen scenarios, we select 88 images from NC4K \cite{lv2021simultaneously} and MoCA-Mask \cite{DBLP:conf/cvpr/ChengXFZHDG22} datasets that are not contained in the 69 categories of camouflaged objects provided by COD10K, such as pangolin, pallas's cat and binturong. As shown in Tab. \ref{tab:tab10}, we compare our HGINet with 12 state-of-the-art COD methods on these unseen scenes. For fair comparison, predictions of the above methods are generated by the models released by authors. It is clear that our HGINet outperforms all methods on these images. Compared to the recently proposed state-of-the-art PUENet \cite{10159663}, our proposed method obtain performance gains of 1.3\%, 4.6\%, 1.6\%, and 5.6\% on for $S_{\alpha} $,  $F_{\beta}^{w} $, $E_{\phi}^{m} $, and $\mathcal{M}$ respectively. Benefiting from the discriminative token excavation and hierarchical feature interaction/fusion of RTFA, HGIT and CAFF, the proposed HGINet achieves great generalization ability in unseen scenes.

\begin{table}[t]
  \footnotesize
  \centering
  \caption{Quantitative comparison with 12 SOTA methods on unseen scenes selected from NC4K and MoCA-Mask datasets. Notes $\uparrow/\downarrow$ denotes the higher/lower the better, and the best and second best are \textbf{bolded} and \underline{underlined} for highlighting, respectively. }
    \begin{tabular}{l|c|cccc}
    \toprule
    \multirow{2}[3]{*}{\textbf{Methods}} & \multirow{2}[3]{*}{\textbf{Year}} & \multicolumn{4}{c}{\textbf{Unseen Scenes (88)}} \\
\cmidrule{3-6} & & $S_{\alpha}\uparrow $ &  $F_{\beta}^{w}\uparrow $  & $E_{\phi}^{m}\uparrow $ & $\mathcal{M}\downarrow$ \\
    \midrule
    SINet\cite{fan2020camouflaged}  & CVPR'20 & 0.878  &   0.818  &  0.919  &  0.038 \\
    MGL-R\cite{zhai2021mutual} & CVPR'21 & 0.866   &  0.786  &  0.891  &  0.037  \\
    UGTR\cite{yang2021uncertainty}  & ICCV'21 & 0.869   &  0.782  &  0.899  &  0.040  \\
    PreyNet\cite{zhang2022preynet} & ACM MM'22 & 0.880   &  0.814  &  0.913  &  0.034  \\
    BGNet\cite{sun2022boundary} & IJCAI'22 & 0.899   &  0.849  &  0.937  &  0.031  \\
    ZoomNet\cite{pang2022zoom} & CVPR'22 & 0.889  &  0.821  &  0.910  &  0.027  \\
    SINet-v2\cite{fan2021concealed} & TPAMI'22 & 0.881  &  0.811  &  0.911  &  0.032  \\
    FEDER\cite{he2023camouflaged} & CVPR'23 & 0.880  &  0.826  &  0.925  &   0.035  \\
    FSPNet\cite{huang2023feature} & CVPR'23 & 0.899   &  0.832  &  0.926  &  0.029  \\
    PUENet\cite{10159663} & TIP'23 &   0.919   &  0.869  &  0.955  &  \underline{0.018} \\
    FSNet\cite{10103836}  & TIP'23 & \underline{0.928}   &  \underline{0.894}  &  \underline{0.965}  &  0.020 \\
    CamoDiff\cite{DBLP:conf/aaai/ChenSL24}  & AAAI'24 &  0.922   &  0.886  &  0.952  &  0.021 \\
    \midrule
    \textbf{HGINet (Ours)} &   \textendash   & \textbf{0.931}  &  \textbf{0.909}  &  \textbf{0.970}  &  \textbf{0.017} \\
    \bottomrule
    \end{tabular}%
  \label{tab:tab10}%
\end{table}%

\subsection{Failure Cases}
\begin{figure}[t]

\centering
\setlength{\abovecaptionskip}{1mm}
\setlength{\belowcaptionskip}{-5mm}
\includegraphics[scale=0.363]{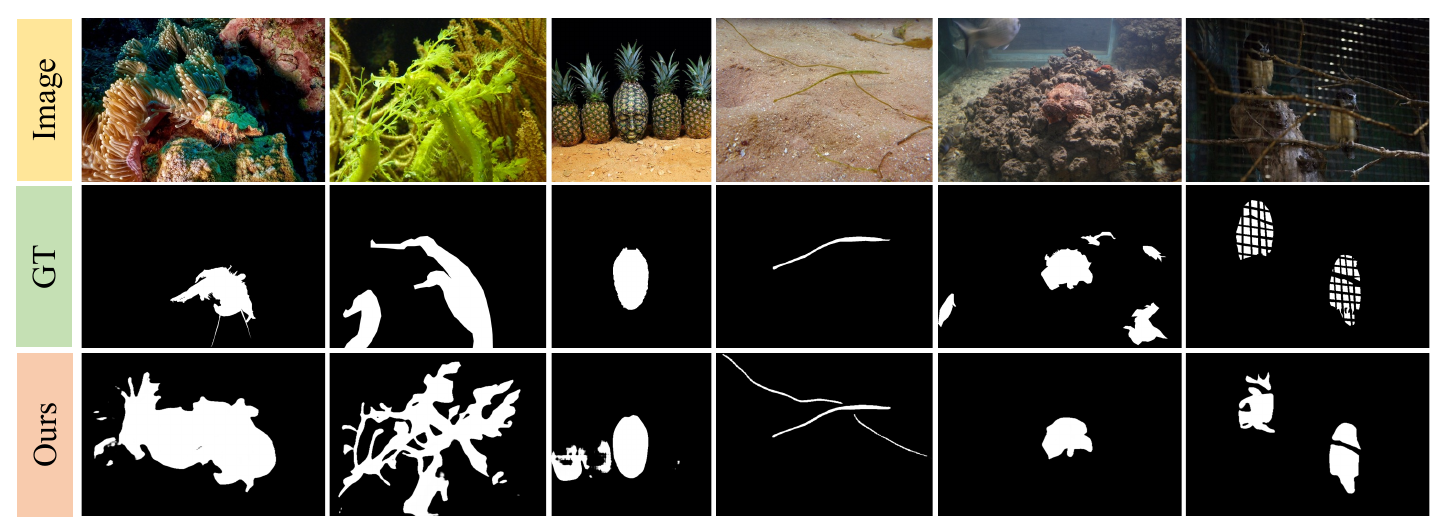}
\caption{Failure cases in several challenging scenarios. Please zoom in for details.}
\label{fig:fig8}
\end{figure}

Although our proposed HGINet achieves state-of-the-art performance, it does not perform well in several challenging scenarios. As shown in Fig. \ref{fig:fig8}, the results in the first four columns indicate that it can not distinguish the camouflaged objects in extremely imperceptible scenes. The last two columns reveal that our method fails to handle multiple or occluded objects in low-light images. These drawbacks will be the improvement directions in our future work.

\section{Conclusions}
In this paper, we propose HGINet for camouflaged object detection, which is capable of discovering imperceptible objects using the hierarchical graph interaction mechanism. To achieve this, a region-aware token focusing attention (RTFA) with dynamic token clustering is designed to excavate the potentially distinguishable tokens in the local region. Then we propose a hierarchical graph interaction transformer (HGIT) to construct bidirectional communication between hierarchical features in the latent interaction space to enhance the visual semantics. Furthermore, we propose a decoder network with confidence aggregated feature fusion (CAFF) modules to refine the local details in ambiguous regions. Extensive experiments conducted on the prevalent COD benchmarks demonstrate that HGINet achieves more competitive performance than other state-of-the-art methods.

%

\bibliographystyle{IEEEtran}

%
%
%
%
%

\vfill

\end{document}